\pdfoutput=1
\documentclass[10pt, a4paper]{article}
\usepackage{lrec-coling2024} 
\usepackage[T1]{fontenc}
\usepackage[utf8]{inputenc}
\usepackage{helvet}

\usepackage{latexsym}
\usepackage{graphicx}
\usepackage{amssymb}
\usepackage{amsmath}
\usepackage{stackrel}
\usepackage{booktabs}
\usepackage{color}
\usepackage{multirow}
\usepackage{listings}
\usepackage{array}
\usepackage{rotating}
\usepackage{CJKutf8}
\usepackage[linesnumbered,ruled,vlined]{algorithm2e}
\usepackage{natbib}
\usepackage{multibib}
\makeatletter
\def\@mb@citenamelist{cite,citep,citet,citealp,citealt,citepalias,citetalias}
\makeatother
\newcites{languageresource}{~}

\usepackage{graphicx}
\usepackage{tabularx}
\usepackage{soul}

\definecolor{dkgreen}{rgb}{0,0.6,0}
\definecolor{gray}{rgb}{0.5,0.5,0.5}
\definecolor{mauve}{rgb}{0.58,0,0.82}

\usepackage{titlesec}
\titleformat{\section}{\normalfont\large\bfseries\center}{\thesection.}{1em}{}
\titleformat{\subsection}{\normalfont\SmallTitleFont\bfseries\raggedright}{\thesubsection.}{1em}{}
\titleformat{\subsubsection}{\normalfont\normalsize\bfseries\raggedright}{\thesubsubsection.}{1em}{}
\renewcommand\thesection{\arabic{section}}
\renewcommand\thesubsection{\thesection.\arabic{subsection}}
\renewcommand\thesubsubsection{\thesubsection.\arabic{subsubsection}}

\usepackage{xcolor}
\usepackage{hyperref}
 \definecolor{darkblue}{rgb}{0, 0, 0.5}
  \hypersetup{colorlinks=true, citecolor=darkblue, linkcolor=darkblue, urlcolor=darkblue}

\usepackage{xstring}

\usepackage{color}

\title{ChatUIE: Exploring Chat-based Unified Information Extraction using Large Language Models}
\name{Jun Xu, Mengshu Sun, Zhiqiang Zhang and Jun Zhou} 

\address{Ant Group, Hangzhou, China \\
         \{xujun.xj, mengshu.sms, lingyao.zzq, jun.zhoujun\}@antgroup.com\\}

\abstract{
Recent advancements in large language models have shown impressive performance in general chat. However, their domain-specific capabilities, particularly in information extraction, have certain limitations. Extracting structured information from natural language that deviates from known schemas or instructions has proven challenging for previous prompt-based methods. This motivated us to explore domain-specific modeling in chat-based language models as a solution for extracting structured information from natural language. In this paper, we present ChatUIE, an innovative unified information extraction framework built upon ChatGLM. Simultaneously, reinforcement learning is employed to improve and align various tasks that involve confusing and limited samples. Furthermore, we integrate generation constraints to address the issue of generating elements that are not present in the input. Our experimental results demonstrate that ChatUIE can significantly improve the performance of information extraction with a slight decrease in chatting ability.
\\ \newline  }

\begin{document}

\maketitleabstract

\section{Introduction}
Information extraction (IE) is a structured prediction task that aims to identify and structure user-specified information from unstructured texts ~\citep{DBLP:conf/anlp/AndersenHWHSN92,DBLP:journals/nle/Grishman19,lu-etal-2022-unified,cao-etal-2022-oneee,jiang-etal-2021-named,DBLP:journals/access/XuS22}. IE tasks are highly diversified due to their varying targets (entities, relations, events, etc.), heterogeneous structures (spans, triplets, records, etc.), and domain-specific schemas~\citep{DBLP:journals/corr/abs-2301-03282,DBLP:journals/corr/abs-2210-02414,DBLP:conf/acl/DuQLDQY022}.
The primary studies~\citep{jiang-etal-2021-named,li2022unified,DBLP:conf/cikm/XuSLF18,ye-etal-2022-packed,cao-etal-2022-oneee,sheng-etal-2021-casee,Zhang2022OptimizingBF,tang-etal-2022-unirel,xu-etal-2022-extracting} of information extraction are task-specialized, which results in dedicated architectures, isolated models, and specialized knowledge sources for different IE tasks. Several improved methods~\citep{lu-etal-2022-unified,DBLP:journals/corr/abs-2301-03282,wei2023zeroshot,wang2023instructuie} have been proposed for the unified modeling of information extraction tasks, including prompt-based extractive and generative models. 
\begin{figure}[htbp]
\centering 
\includegraphics[width=7.8cm]{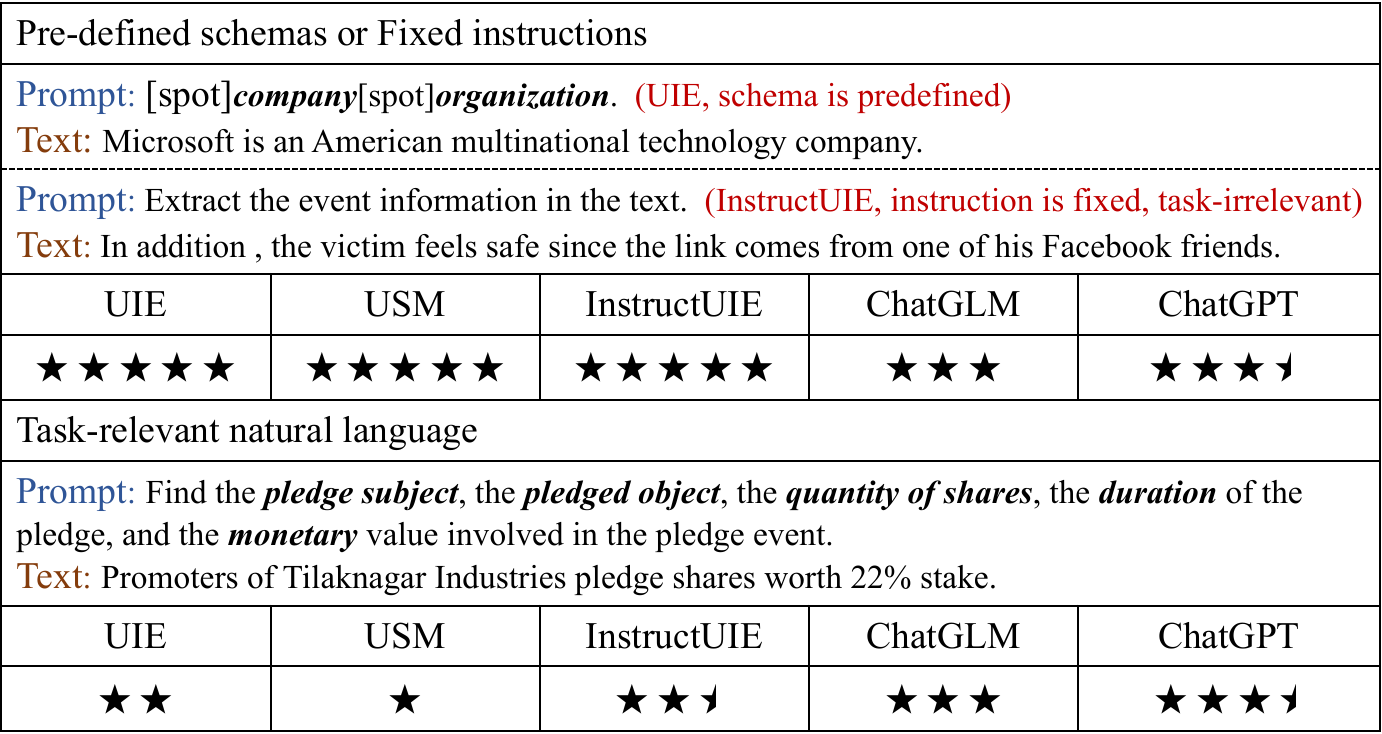} 
\caption{The approximate performance of unified information extraction framework in various application scenarios.} 
\label{Fig.motivation}
\end{figure}
However, these methods are highly tailored to pre-defined schemas or fixed instructions, which makes it extremely challenging to facilitate natural language extraction. 
As shown in Figure~\ref{Fig.motivation}, UIE relies on a pre-defined schema and prompt template. Deviating from this consistency can significantly degrade model performance, especially for zero-shot tasks where the schema was not seen during training. In contrast, InstructUIE uses a set of instructions for information extraction. However, since these instructions are not tailored to the specific task (what to extract?), the model is restricted to the known dataset. When faced with a new schema, using task-irrelevant instructions makes it difficult to produce satisfactory results.

Generally, previous instruction-based methods focused more on memorizing instructions rather than comprehending them. While ChatGLM outperforms InstructUIE in task-relevant natural language scenarios, there is room for improvement in closed domain. 
However, enhancing the information extraction capabilities while preserving the general chat capabilities of ChatGLM presents challenges. Firstly, conflicts between domain-specific knowledge and the knowledge embedded in LLMs may result in  knowledge forgetting. Secondly, the scarcity of annotated domain knowledge and uneven sample distribution make it arduous for LLM to effectively adapt and accommodate.
These issues cannot be adequately addressed through supervised fine-tuning alone. In order to address this issue, we have introduced reinforcement learning to align various tasks. In contrast to other tasks, the result for information extraction is derived from the input. To ensure that the generated elements remain within the input, we utilize generation constraint decoding.
\begin{figure}[htbp]
\centering 
\includegraphics[width=7.7cm]{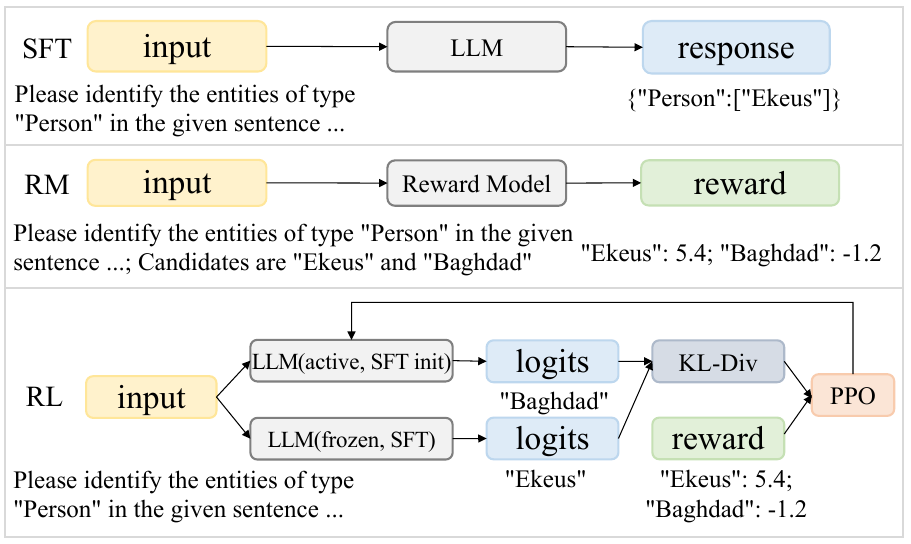} 
\caption{The overall framework of chat-based unified information extraction using LLMs.} 
\label{Fig.overview_detail} 
\end{figure}

\section{Methodology}
The architecture of our framework is illustrated in Figure~\ref{Fig.overview_detail}, which mainly consists of three stages. Initially, supervised fine-tuning is used to incorporate domain knowledge into LLMs. Next, the reward learning model is utilized to enhance the learning of confusing samples and data with limited samples. Lastly, we combine the trained supervised fine-tuning model and reward model using reinforcement learning to align various tasks.

\subsection{Domain Knowledge Integration}
\begin{figure}[htbp]
\centering 
\includegraphics[width=6.0cm]{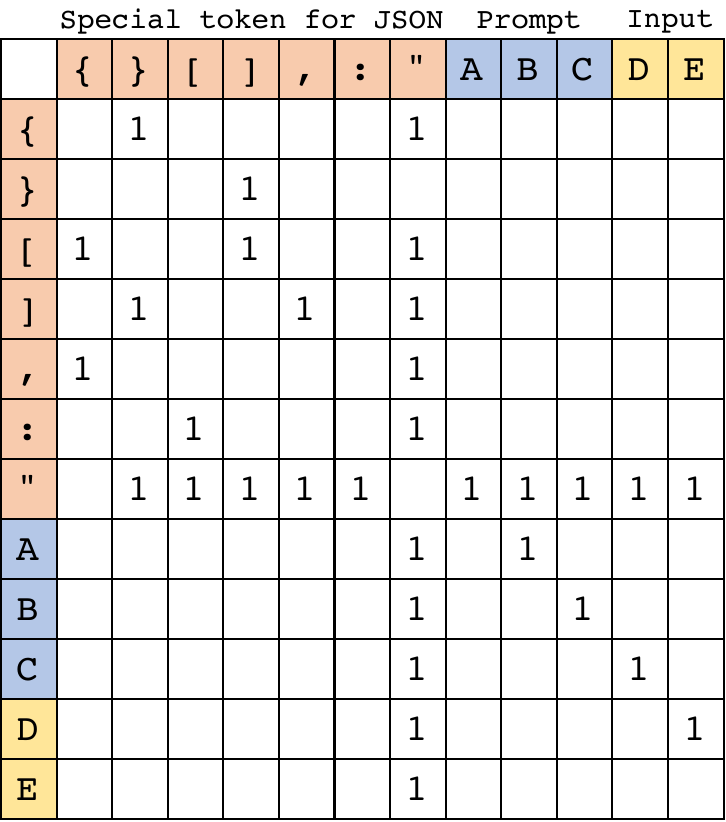} 
\caption{Strategies for generating constraints in information extraction tasks.} 
\label{Fig.generating_constraints} 
\end{figure}

We utilize supervised learning to fine-tune ChatGLM using a variety of information extraction datasets. 
The input of the SFT model is divided into two parts: instruction and context. The instruction, with $M$ tokens, and the context, with $N$ tokens, are encoded by GLM to derive vector representations $x=[x_{i1},...,x_{iM};x_{c1},...,x_{cN}] \in \mathbb{R}^{(M+N)\times D}$, where $D$ represents the dimension of the embedding.
The probability of generating each token is as follows:
\begin{equation}
\begin{aligned}
p(y_i|x) = \frac{p(v(y_i)|c(x;y_{i-1}))}{\sum_{y'\in\mathcal{V}}{p(v(y')|c(x;y_{i-1}))}}
\end{aligned}
\label{sft_prob}
\end{equation}
where $\mathcal{V}$ represents a vocabulary of 130, 528 tokens. $v$ is a MLP , and $c$ is the decoder of GLM. The dimension of $p(y_i|x)$ is $\mathbb{R}^{|\mathcal{V}|}$. 
The objective function of SFT is to maximize the likelihood:
\begin{equation}
\begin{aligned}
\mathcal{L}_{SFT} = \frac{1}{L}\sum_{i=0}^{L}{CE(y_i,p(y_i|x))}
\end{aligned}
\label{sft_loss}
\end{equation}
where $L$ represents the target sequence length. CE is the cross-entropy loss. Unified information extraction stands out from other text generation tasks as it necessitates the generated content to be a span within the input. Moreover, our framework adopts a JSON format to effectively represent the structured relationships. Therefore, we introduce generative constraint decoding. As shown in Figure \ref{Fig.generating_constraints}, for instance, after generating token `A', the next token must be either `B' or `"'.

\subsection{Reinforcement Learning}
In unified information extraction, the presence of diverse data sources and types often leads to challenges, such as type confusion and uneven distribution of samples, in the supervised fine-tuning model. To address these issues, reinforcement learning is introduced as a solution. 
The reward model of reinforcement learning also uses ChatGLM as the backbone.
It takes an instruction, a context, and a positive or negative response as input, and outputs a scalar reward.
We use the logits of the EOS token to represent the scalar reward:
\begin{equation}
\begin{aligned}
r(x,y)=v(y_{eos}|c(x;y))
\end{aligned}
\label{rm_positive}
\end{equation}
Our goal is to maximize the difference between the rewards of positive and negative samples. Therefore, the objective of reward modeling can be expressed as follows:
\begin{equation}
\begin{aligned}
\mathcal{L}_{RM} = -log(\sigma(r(x_p,y_p)-r(x_n,y_n)))
\end{aligned}
\label{rm_loss}
\end{equation}
where $\sigma$ is a sigmoid function. 
Contrary to the supervised fine-tuning model, the reward model does not exclusively dependent on training samples from the training set. The construction of training samples encompasses diverse methods, including: (1) substituting sample results with different types of confusion to generate negative samples, and (2) using the SFT and ChatGPT models to forecast extraction results for external analogous datasets. Moreover, ChatGPT is used to score the extraction results, thereby increasing the amount of limited sample data.
Then, the optimization strategy for reinforcement learning adopts PPO. 
Finally, the objective function in RL training can be expressed as follows:
\begin{equation}
\begin{aligned}
\mathcal{L}_{RL} = r(x,y)-\beta log(\frac{p^{RL}(y|x)}{p^{SFT}(y|x)})
\end{aligned}
\label{p_loss}
\end{equation}
where $r(x,y)$ represents a scalar reward, and $\beta$ is a scalar coefficient of KL-div. $p^{RL}(y|x)$ denotes the logits of the active model, while $p^{SFT}(y|x)$ denotes the logits of the reference model.

\section{Experimental Settings and Results}
\subsection{Experimental Settings}

\noindent\textbf{{Dataset}}
We conducted our experiments on multiple widely-used datasets, including Resume for NER; CoNLL-2004 for RE; FewFC for EE~\citep{zhang-yang-2018-chinese,roth-yih-2004-linear,Yang2021}; WebQA, CEval, CMMLU, and MMLU ~\citep{DBLP:journals/corr/LiLHWCZX16,huang2023ceval,li2023cmmlu,hendryckstest2021} for general chat. For detailed information on data division, please refer to Table~\ref{tab:datasets}. The training of ChatUIE requires two sets of data: one for the reward model (<instruction, context, positive response, negative response>) and the other for supervised fine-tuning and reinforcement learning (<instruction, context, response>).
\begingroup
\renewcommand\arraystretch{1.2}
\begin{table}[htbp]
\centering
\small
\setlength\aboverulesep{0pt}\setlength\belowrulesep{0pt}
\begin{tabular}{p{0.4cm}<\centering p{0.9cm}<\centering p{1.1cm}<\centering p{1.1cm}<\centering p{1.1cm}<\centering}
\toprule
                         &          & SFT & RM & RL \\ \hline
\multirow{4}{*}{\rotatebox{90}{Train}}    & Resume & 10,472     & 5,489        & 10,472                 \\ \cline{2-5}
                               & CoNLL    & 2,647 & 1,153    & 2,647       \\ \cline{2-5}
                               & FewFC     & 25,665    & 8,299 & 25,665  \\ \cline{2-5}
                               & Sum     & \textbf{38,784}    & \textbf{14,941} & \textbf{38,784}  \\\midrule
\multirow{4}{*}{\rotatebox{90}{Dev}} & Resume & 1,258  & 657  & 1,258   \\ \cline{2-5} 
                         & CoNLL    & 659  & 288  & 659    \\ \cline{2-5} 
                         & FewFC     & 3,263  & 1,050  & 3,263   \\ \cline{2-5}
                               & Sum     & \textbf{5,180}    & \textbf{1,996} & \textbf{5,180}  \\\midrule
\multirow{4}{*}{\rotatebox{90}{Test}}  & Resume & 1,253  & 664  & 1,253   \\ \cline{2-5} 
                        & CoNLL    & 641  & 288  & 641    \\ \cline{2-5} 
                         & FewFC     & 3,244  & 1,049   & 3,244  \\ \cline{2-5}
                               & Sum     & \textbf{5,138}    & \textbf{2,001} & \textbf{5,138}  \\ \bottomrule
\end{tabular}
\caption{Details of the datasets: The division of these datasets is consistent with previous work, and ChatUIE performed training/dev data augmentation based on the divided results.}
\label{tab:datasets}
\end{table}
\endgroup
Negative responses consist of confusing data or data that the SFT model cannot fit. For example, an instruction like ``Please identify the entities of type Person in the given sentence'' and a context like ``U.N. official Ekeus heads for Baghdad''. The training data for SFT and RL is constructed as follows: 
\begin{lstlisting}
{
	"instruction": "Please identify the entities of type Person in the given sentence",
	"context": "U.N. official Ekeus heads for Baghdad",
	"ouput": '[{"Person": ["EKeus"]}]'
}
\end{lstlisting}
The training data for RM is constructed as follows:

\begin{lstlisting}
{
	"instruction": "Please identify the entities of type `Person' in the given sentence",
	"context": "U.N. official Ekeus heads for Baghdad",
	"ouput": [
        '[{"Person": ["EKeus"]}]', 
        '[{"Person": ["Baghdad"]}]'
    ]
}
\end{lstlisting}
\noindent\textbf{{Evaluation Metrics.}}
For the NER task, we follow a span-level evaluation setting, where the entity span and entity type must be correctly predicted. For the RE task, a relation triple is correct if the model correctly predicts the span of subject and object and the relation between subject and object. For the EE task, we report two evaluation metrics: (1) Event Trigger: an event trigger is correct if the event type and the trigger span are correctly predicted. (2) Event Argument: an event argument is correct if its role type and event type match a reference argument mention. For general chat task, we use ROUGE-1 as the metric, it refers to the overlap of unigrams between the generation and reference response.



\noindent\textbf{{Implementation Details.}} 
We compare the proposed ChatUIE with the following strong baseline models: 
\textbf{ChatGLM} \footnote{https://github.com/THUDM/ChatGLM-6B} takes information extraction as a generation problem. The input and output of the test set are consistent with ChatUIE, but ChatGLM has not been trained, and the output is inferred directly. \textbf{ChatGPT} gets responses through the official interface. \textbf{UIE} implementation and parameters are consistent with the author's official code \footnote{https://github.com/universal-ie/UIE}, the English dataset uses the UIE English base model \footnote{https://huggingface.co/luyaojie/uie-base-en}, and the Chinese dataset uses the mT5 base model. 
Our model \textbf{ChatUIE} is trained on 8$\times$V100-32G. For other hyper-parameters and details, please refer to Table \ref{tab:hyperparameters}.
\begingroup
\renewcommand\arraystretch{1.1}
\begin{table}[htbp]
\centering
\small
\setlength\aboverulesep{0pt}\setlength\belowrulesep{0pt}
\begin{tabular}{p{2.7cm}|p{0.85cm}<\centering p{0.85cm}<\centering p{0.85cm}<\centering}
\toprule
                     & SFT    & RM     & RL      \\ \hline
batch size           & 64     & 32     & 8       \\ \hline
fine-tuning type     & LoRA   & LoRA   & LoRA    \\ \hline
train epochs         & 0-15   & 3      & 2       \\ \hline
lora rank            & 8      & 8      & 8       \\ \hline
lora dropout         & 0.1    & 0.1    & 0.1     \\ \hline
lora target          & QKV    & QKV    & QKV     \\ \hline
learning rate        & 1e-4   & 2e-5   & 1e-6    \\ \hline
max input length     & 450    & 450    & 450     \\ \hline
max output length    & 600    & 600    & 600     \\ \hline
KL-div $\beta$       & -      & -      & 0.1     \\ \bottomrule 
\end{tabular}
\caption{Hyperparameters of different models.}
\label{tab:hyperparameters}
\end{table}
\endgroup
The implementation of supervised fine-tuning and reward modeling refers to ChatGLM~\citep{DBLP:conf/acl/DuQLDQY022,DBLP:conf/iclr/RobinsonW23}, and the implementation of reinforcement learning refers to TRL~\citep{vonwerra2022trl}. The results reported in the experiment are the average of 5 different random seeds (0,1,2,3,4).

\subsection{Results}
\subsubsection{Overall Results}
\begin{figure}[htbp!]
\centering 
\includegraphics[width=7.5cm]{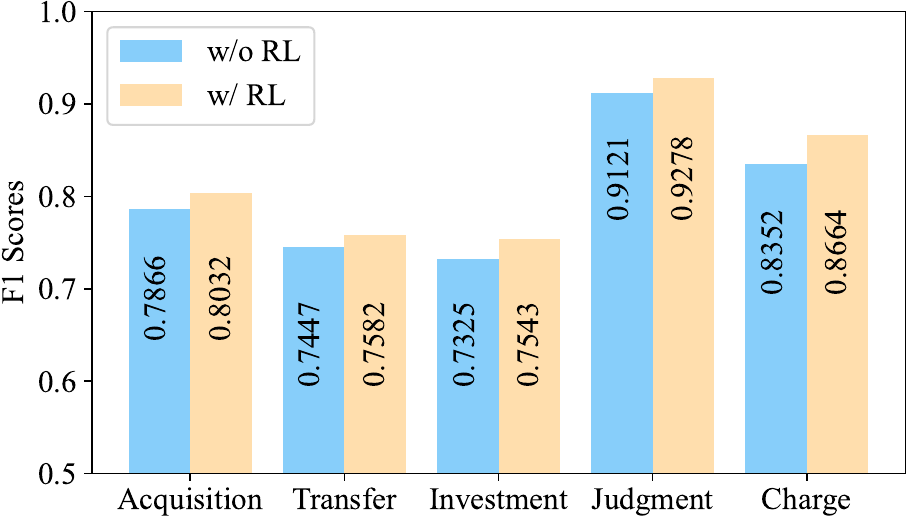} 
\caption{Performance comparison of confusing and limited samples after reinforcement learning in FewFC.} 
\label{Fig.event_type_result} 
\end{figure}

The results reported in the experiment are the average of five different random seeds. Table~\ref{tab:overall_ner_re} and Table~\ref{tab:overall_event_extraction} present the comparisons between our model and other baselines. As demonstrated, our model surpassed the generative model UIE (with supervised fine-tuning) by 3.89\%, 1.27\%, and 5.75\% in F1 score on the Resume, CoNLL, and FewFC datasets, respectively. Additionally, it can be observed that the performance of ChatGLM and ChatGPT has significantly decreased without domain-specific data training. 
After applying reinforcement learning (RL), ChatUIE exhibited performance improvements of 1.56\% in Resume, 2.43\% in CoNLL, and 3.59\% in FewFC. Furthermore, the generation constraints (GC) also showed noticeable enhancements across different datasets.
As shown in Figure \ref{Fig.event_type_result}, using FewFC as an example, the categories of Acquisition, Transfer, and Investment are susceptible to confusion. However, after applying reinforcement learning techniques, the performance of all three types  improves significantly. Notably, the sample sizes for Judgment and Charge are relatively small, constituting less than one-third of the Investment samples. By incorporating external homologous data through reinforcement learning, the overall effect improves by approximately 1 to 3 percentage points.
\begingroup
\renewcommand\arraystretch{1.1}
\begin{table}[htbp]
\centering
\small
\setlength\aboverulesep{0pt}\setlength\belowrulesep{0pt}
\begin{tabular}{p{1.15cm} p{0.52cm}<\centering p{0.52cm}<\centering p{0.52cm}<\centering p{0.52cm}<\centering p{0.52cm}<\centering p{0.52cm}<\centering}
\toprule
 \multicolumn{1}{l}{\multirow{2}{*}{Model}} & \multicolumn{3}{c}{Resume (\%)} & \multicolumn{3}{c}{CoNLL (\%)} \\ \cline{2-7} 
 \multicolumn{1}{l}{}   & P     & R     & F1   & P     & R     & F1     \\ \hline
ChatGLM  &  41.05  &  86.98  &  55.78  &  11.60  &  70.82  &  19.93 \\ \hline
ChatGPT  &  80.47  &  63.49  &  70.98  &  48.03  &  24.93  &  32.83 \\ \hline
UIE  &  94.03  &  93.67  &  93.85  &  75.56  &  73.72  &  74.63 \\ \hline
ChatUIE  &  95.58  &  95.82  &  \textbf{95.70}  &  75.82  &  75.35  &  \textbf{75.58} \\ \hline
\,\,w/o RL &  93.51  &  94.97  &  94.23  &  74.25  &  73.34  &  73.79 \\ \hline
\,\,w/o GC &  94.37  &  94.55  &  94.46  &  74.03  &  73.82  &  73.92 \\ 
\bottomrule
\end{tabular}
\caption{Overall results of Resume and CoNLL.}
\label{tab:overall_ner_re}
\end{table}
\endgroup
\begingroup
\renewcommand\arraystretch{1.1}
\begin{table}[htbp]
\centering
\small
\setlength\aboverulesep{0pt}\setlength\belowrulesep{0pt}
\begin{tabular}{p{1.15cm} p{0.52cm}<\centering p{0.52cm}<\centering p{0.52cm}<\centering p{0.52cm}<\centering p{0.52cm}<\centering p{0.52cm}<\centering}
\toprule
 \multicolumn{1}{l}{\multirow{2}{*}{Model}} & \multicolumn{3}{c}{TC (\%)} & \multicolumn{3}{c}{AC (\%)} \\ \cline{2-7} 
 \multicolumn{1}{l}{}   & P     & R     & F1   & P     & R     & F1     \\ \hline
ChatGLM  &  20.33  &  75.87  &  32.08  &  10.33  &  61.21  &  17.68 \\ \hline
ChatGPT  &  83.86  &  42.56  &  56.47  &  76.80  &  29.54  &  42.67 \\ \hline
UIE  &  89.64  &  76.90  &  82.78  &  78.82  &  61.42  &  69.04 \\ \hline
ChatUIE  &  87.04  &  79.69  &  \textbf{83.20}  &  73.63  &  72.39  &  \textbf{73.01} \\ \hline
\,\,w/o RL  &  86.58  &  76.81  &  81.40  &  78.53  &  63.93  &  70.48 \\ \hline
\,\,w/o GC  &  86.84  &  77.92  &  82.14  &  76.78  &  66.56  &  71.31 \\ \hline
\bottomrule
\end{tabular}
\caption{Overall results of event extraction on FewFC. 
 TC is trigger classification, while AC stands for argument classification.
}
\label{tab:overall_event_extraction}
\end{table}
\endgroup

\subsubsection{Results For Chatting}
\begin{figure}[htbp]
\centering 
\includegraphics[width=8.0cm]{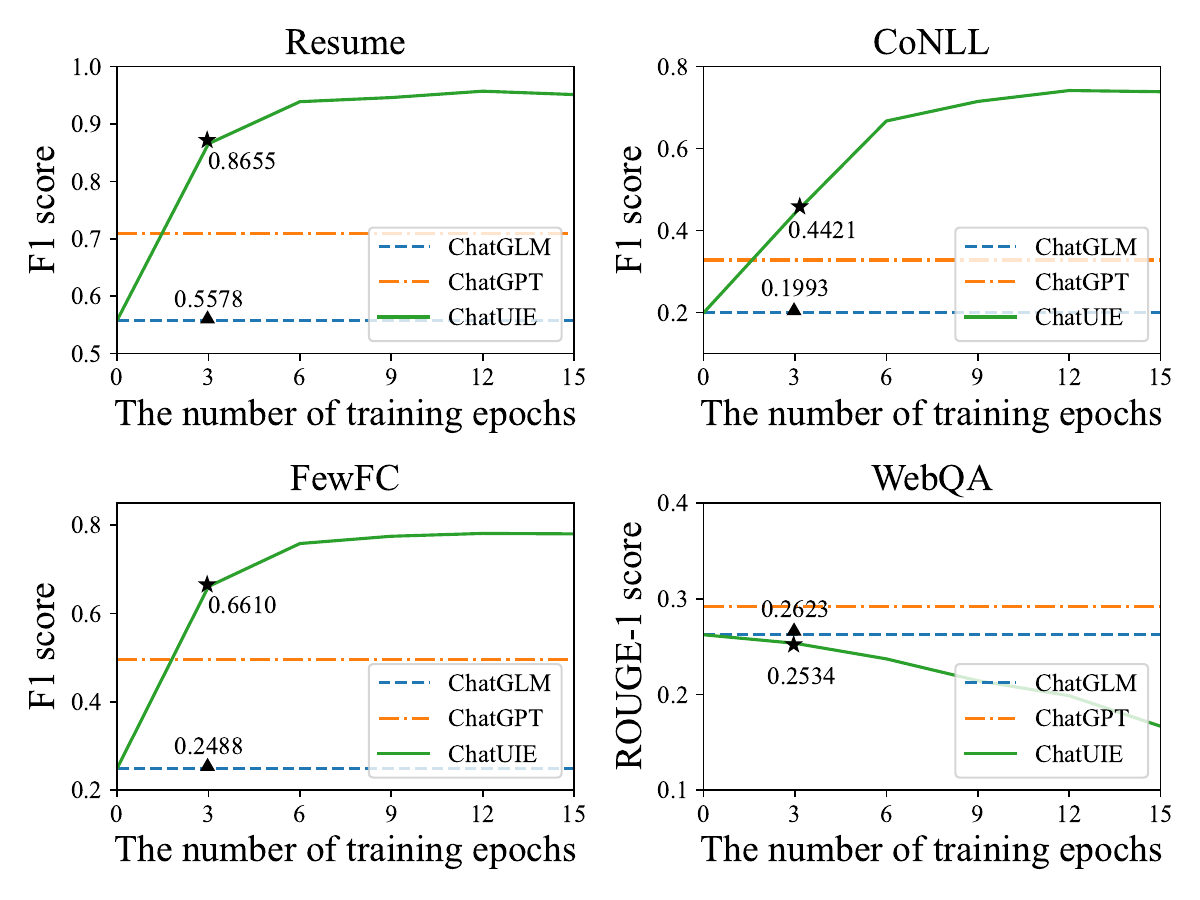} 
\caption{Performance of different tasks as the number of training epochs increase.} 
\label{Fig.domain_detail} 
\vspace{-0.1cm}
\end{figure}

To assess the overall chat capability of ChatUIE, we have chosen WebQA as the dataset for evaluating common sense question and answer performance. As shown in Figure~\ref{Fig.domain_detail}, compared to ChatGLM, ChatUIE demonstrated a significant improvement of 30.77\% on Resume, 24.28\% on CoNLL, and 41.22\% when the epoch is 3. However, there is only a slight decrease of 0.89\% in the general question answering task, which is still within an acceptable range. 
As the number of training epochs increases, the performance of the domain-specific datasets improves. However, there is a risk of losing the general chatting ability. Simultaneously, we also evaluate the overall capabilities of ChatUIE using a dataset for large-scale model assessment. As depicted in Table \ref{tab:overall_ceval_mmlu} and Table \ref{tab:overall_cmmlu}, a minor decline is observed in the overall chat capabilities of ChatUIE. At the same time, it is evident that the absence of reinforcement learning in ChatUIE results in a partial decrease in its overall chat capabilities. Reinforcement learning effectively mitigates knowledge decay by integrating diverse tasks.
\begingroup
\renewcommand\arraystretch{1.1}
\begin{table}[htbp]
\centering
\small
\setlength\aboverulesep{0pt}\setlength\belowrulesep{0pt}
\begin{tabular}{p{0.1cm}<\centering p{1.1cm} p{0.7cm}<\centering p{0.7cm}<\centering p{0.7cm}<\centering p{0.7cm}<\centering p{0.7cm}<\centering}
\toprule
&& STEM & Hum & SSci & Other & Avg \\ \hline
\multirow{3}{*}{\rotatebox{90}{\footnotesize CEval }}&ChatGLM & 35.84 & 44.24 & 45.14 & 40.07 & 40.30 \\ \cline{2-7}
&ChatUIE & 35.12 & 43.93 & 45.02 & 39.46 & 39.72 \\ \cline{2-7}
&- RL & 34.89 & 43.64 & 44.37 & 39.05 & 39.33 \\ \bottomrule
\multirow{3}{*}{\rotatebox{90}{\footnotesize MMLU }}&ChatGLM & 34.93 & 43.03 & 45.40 & 42.47 & 40.83 \\  \cline{2-7}
&ChatUIE & 35.02 & 43.22 & 44.17 & 41.23 & 40.25 \\ \cline{2-7}
&- RL & 34.85 & 43.11 & 43.83 & 40.96 & 40.02 \\ \bottomrule
\end{tabular}
\caption{Overall results of CEval and MMLU.}
\label{tab:overall_ceval_mmlu}
\end{table}
\endgroup


\begingroup
\renewcommand\arraystretch{1.1}
\begin{table}[htbp]
\centering
\small
\setlength\aboverulesep{0pt}\setlength\belowrulesep{0pt}
\begin{tabular}{p{1.2cm} p{0.55cm}<\centering p{0.55cm}<\centering p{0.55cm}<\centering p{0.55cm}<\centering p{0.55cm}<\centering p{0.55cm}<\centering}
\toprule
& STEM & Hum & SSci & Other & CSp & Avg \\ \hline
ChatGLM & 31.53 & 40.52 & 41.30 & 39.88 & 38.59 & 38.35 \\ \hline
ChatUIE & 30.97 & 40.23 & 40.88 & 38.24 & 39.12 & 37.86 \\ \hline
- RL & 30.25 & 40.34 & 40.37 & 38.02 & 38.78 & 37.49 \\ \bottomrule
\end{tabular}
\caption{Overall results of CMMLU.}
\label{tab:overall_cmmlu}
\end{table}
\endgroup


\subsubsection{Zero-Shot Information Extraction}
We evaluate the zero-shot performance of ChatUIE by testing it on unseen information extraction datasets. These include MSRA~\citep{levow-2006-third} for NER
; SemEval~\citep{hendrickx-etal-2010-semeval} for RE
; and iFLYTEK for EE. As can been seen from Table~\ref{tab:overall_zeroshot_ner_re} and Table~\ref{tab:overall_zeroshot_event_extraction}, our model outperformed the baseline model (ChatGLM) by 18.85\%, 9.58\%, and 8.89\% in F1 score on the MSRA, SemEval, and iFLYTEK datasets, respectively. Since the training dataset of InstructUIE includes SemEval, zero-shot testing is not conducted. ChatUIE surpasses UIE and InstructUIE as it doesn't rely on pre-defined schema or fixed instructions, enabling it to comprehend natural language more effectively.
\begingroup
\renewcommand\arraystretch{1.1}
\begin{table}[htbp!]
\centering
\small
\setlength\aboverulesep{0pt}\setlength\belowrulesep{0pt}
\begin{tabular}{p{1.2cm}<\centering p{0.55cm}<\centering p{0.55cm}<\centering p{0.55cm}<\centering p{0.55cm}<\centering p{0.55cm}<\centering p{0.55cm}<\centering}
\toprule
 \multicolumn{1}{c}{\multirow{2}{*}{Model}} & \multicolumn{3}{c}{MSRA (\%)} & \multicolumn{3}{c}{SemEval (\%)} \\ \cline{2-7} 
 \multicolumn{1}{c}{}   & P     & R     & F1   & P     & R     & F1     \\ \hline
ChatGLM  &  14.74  &  88.31  &  25.26  &  6.49  &  32.84  &  10.84 \\ \hline
ChatGPT  &  59.63  &  28.73  &  38.78  &  58.53  &  9.21  &  15.92 \\ \hline
InstructUIE	     &  29.35  &  33.56	 &  31.31  &  -	& -	& -  \\ \hline
UIE	     &  24.33  &  40.97	 &  30.52  &  9.87	& 47.26	& 16.33  \\ \hline
ChatUIE  &  30.73  &  78.14  &  \textbf{44.11}  &  15.21  &  31.03  &  \textbf{20.42} \\ \bottomrule
\end{tabular}
\caption{Overall results of zero-shot NER on MSRA and RE on SemEval.}
\label{tab:overall_zeroshot_ner_re}
\end{table}
\endgroup
\begingroup
\renewcommand\arraystretch{1.1}
\begin{table}[htbp!]
\centering
\small
\setlength\aboverulesep{0pt}\setlength\belowrulesep{0pt}
\begin{tabular}{p{1.2cm}<\centering p{0.55cm}<\centering p{0.55cm}<\centering p{0.55cm}<\centering p{0.55cm}<\centering p{0.55cm}<\centering p{0.55cm}<\centering}
\toprule
 \multicolumn{1}{c}{\multirow{2}{*}{Model}} & \multicolumn{3}{c}{TC (\%)} & \multicolumn{3}{c}{AC (\%)} \\ \cline{2-7} 
 \multicolumn{1}{c}{}   & P     & R     & F1   & P     & R     & F1     \\ \hline
ChatGLM  &  9.79  &  37.97  &  15.57  &  8.21  &  27.41  &  12.63 \\ \hline
ChatGPT  &  38.50  &  11.66  &  17.90  &  33.25  &  10.04  &  15.42 \\ \hline
InstructUIE	     & 13.39  &  16.78	&  14.89 &	10.55 & 12.83 &	11.58 \\ \hline
UIE	     &  11.39  &  31.78	&  16.77 &	10.37 & 22.72 &	14.24  \\ \hline
ChatUIE  &  22.82  &  26.41  &  \textbf{24.49}  &  20.07  &  23.25  &  \textbf{21.54} \\ \bottomrule
\end{tabular}
\caption{Overall results of zero-shot event extraction on iFLYTEK.}
\label{tab:overall_zeroshot_event_extraction}
\end{table}
\endgroup

\subsubsection{Case Study}
\begin{figure}[htbp!]
\centering 
\includegraphics[width=7.8cm]{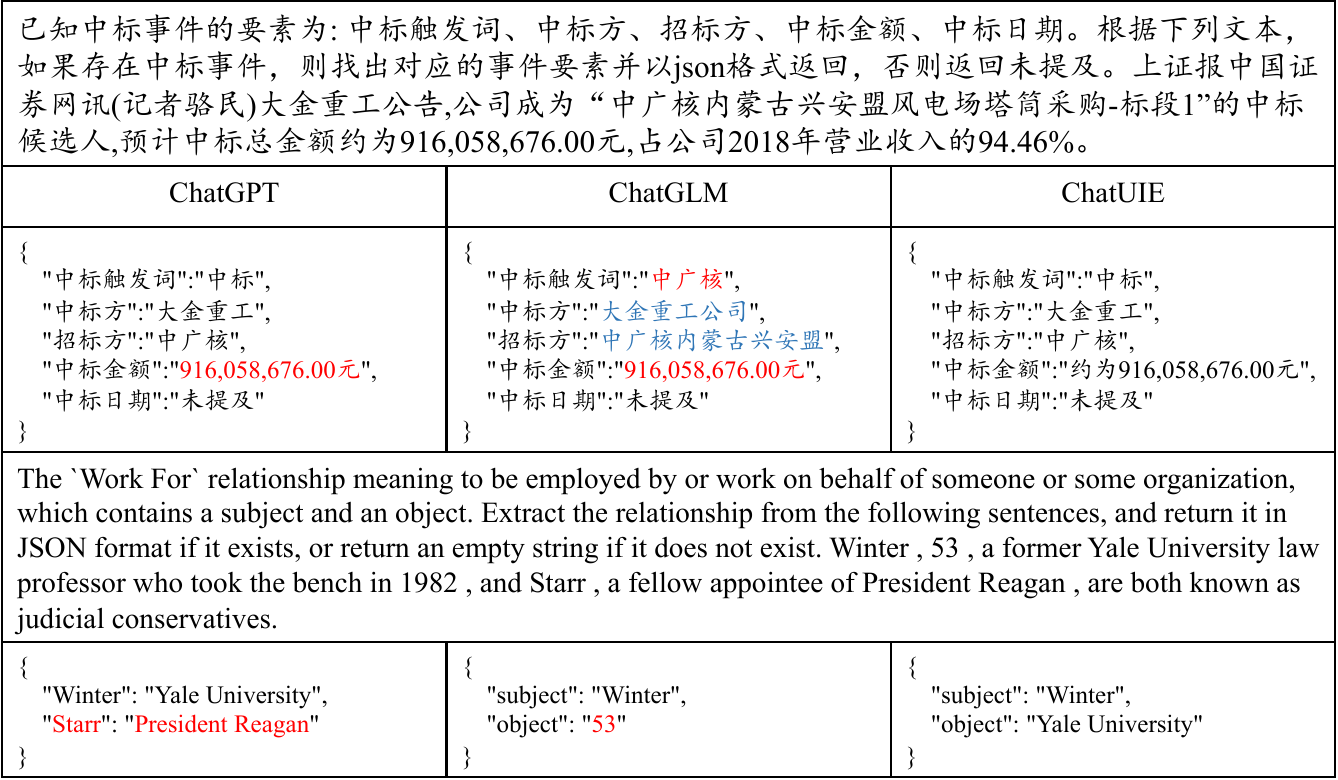} 
\caption{The impacts of various chat models on event extraction and relation extraction.} 
\label{Fig.case_study} 
\end{figure}

As shown in Figure~\ref{Fig.case_study}, it is clear that ChatGPT does not strictly follow the specified format, and some of the generated content in ChatGLM does not match the input. However, ChatUIE addresses these issues by utilizing reinforcement learning and generation constraints. 

\section{Conclusion}
We have presented ChatUIE, a chat-like unified information extraction framework based on ChatGLM. Our framework effectively improves the performance of ChatGLM on domain-specific datasets while preserving its ability to chat. Empirical comparisons and analytical experiments have verified its effectiveness. Moreover, our work may have implications for other complex structured generation tasks.

\section{Limitations}
Nonetheless, these results must be interpreted with caution and several limitations should be borne in mind. 
Firstly, due to GPU limitations, we only trained our model based on ChatGLM-6B. Secondly, the processing speed of generative information extraction is significantly slower than that of extractive information extraction, making it more suitable for interactive applications. Thirdly, while the performance on domain-specific datasets is improved, there may be a slight loss of the ability to chat.

\section{Acknowledgements}
We would like to thank the anonymous reviewers for their constructive comments. This work was
supported by Ant Group.

\section{Bibliographical References}
\bibliographystyle{lrec-coling2024-natbib}
\bibliography{lrec-coling2024-example}

\begin{thebibliography}{29}
\expandafter\ifx\csname natexlab\endcsname\relax\def\natexlab#1{#1}\fi

\bibitem[{Andersen et~al.(1992)Andersen, Hayes, Weinstein, Huettner, Schmandt, and Nirenburg}]{DBLP:conf/anlp/AndersenHWHSN92}
Peggy~M. Andersen, Philip~J. Hayes, Steven~P. Weinstein, Alison~K. Huettner, Linda~M. Schmandt, and Irene~B. Nirenburg. 1992.
\newblock \href {https://doi.org/10.3115/974499.974531} {Automatic extraction of facts from press releases to generate news stories}.
\newblock In \emph{3rd Applied Natural Language Processing Conference, {ANLP} 1992, Trento, Italy, March 31 - April 3, 1992}, pages 170--177. {ACL}.

\bibitem[{Cao et~al.(2022)Cao, Li, Su, Li, Fei, Wu, Li, Zhao, and Ji}]{cao-etal-2022-oneee}
Hu~Cao, Jingye Li, Fangfang Su, Fei Li, Hao Fei, Shengqiong Wu, Bobo Li, Liang Zhao, and Donghong Ji. 2022.
\newblock \href {https://aclanthology.org/2022.coling-1.170} {{O}ne{EE}: A one-stage framework for fast overlapping and nested event extraction}.
\newblock In \emph{Proceedings of the 29th International Conference on Computational Linguistics}, pages 1953--1964, Gyeongju, Republic of Korea. International Committee on Computational Linguistics.

\bibitem[{Du et~al.(2022)Du, Qian, Liu, Ding, Qiu, Yang, and Tang}]{DBLP:conf/acl/DuQLDQY022}
Zhengxiao Du, Yujie Qian, Xiao Liu, Ming Ding, Jiezhong Qiu, Zhilin Yang, and Jie Tang. 2022.
\newblock {GLM:} general language model pretraining with autoregressive blank infilling.
\newblock pages 320--335.

\bibitem[{Grishman(2019)}]{DBLP:journals/nle/Grishman19}
Ralph Grishman. 2019.
\newblock \href {https://doi.org/10.1017/S1351324919000512} {Twenty-five years of information extraction}.
\newblock \emph{Nat. Lang. Eng.}, 25(6):677--692.

\bibitem[{Hendrickx et~al.(2010)Hendrickx, Kim, Kozareva, Nakov, {\'O}~S{\'e}aghdha, Pad{\'o}, Pennacchiotti, Romano, and Szpakowicz}]{hendrickx-etal-2010-semeval}
Iris Hendrickx, Su~Nam Kim, Zornitsa Kozareva, Preslav Nakov, Diarmuid {\'O}~S{\'e}aghdha, Sebastian Pad{\'o}, Marco Pennacchiotti, Lorenza Romano, and Stan Szpakowicz. 2010.
\newblock \href {https://aclanthology.org/S10-1006} {{S}em{E}val-2010 task 8: Multi-way classification of semantic relations between pairs of nominals}.
\newblock In \emph{Proceedings of the 5th International Workshop on Semantic Evaluation}, pages 33--38, Uppsala, Sweden. Association for Computational Linguistics.

\bibitem[{Hendrycks et~al.(2021)Hendrycks, Burns, Basart, Zou, Mazeika, Song, and Steinhardt}]{hendryckstest2021}
Dan Hendrycks, Collin Burns, Steven Basart, Andy Zou, Mantas Mazeika, Dawn Song, and Jacob Steinhardt. 2021.
\newblock Measuring massive multitask language understanding.
\newblock \emph{Proceedings of the International Conference on Learning Representations (ICLR)}.

\bibitem[{Huang et~al.(2023)Huang, Bai, Zhu, Zhang, Zhang, Su, Liu, Lv, Zhang, Lei, Fu, Sun, and He}]{huang2023ceval}
Yuzhen Huang, Yuzhuo Bai, Zhihao Zhu, Junlei Zhang, Jinghan Zhang, Tangjun Su, Junteng Liu, Chuancheng Lv, Yikai Zhang, Jiayi Lei, Yao Fu, Maosong Sun, and Junxian He. 2023.
\newblock C-eval: A multi-level multi-discipline chinese evaluation suite for foundation models.
\newblock \emph{arXiv preprint arXiv:2305.08322}.

\bibitem[{Jiang et~al.(2021)Jiang, Zhang, Cao, Yin, and Zhao}]{jiang-etal-2021-named}
Haoming Jiang, Danqing Zhang, Tianyu Cao, Bing Yin, and Tuo Zhao. 2021.
\newblock \href {https://doi.org/10.18653/v1/2021.acl-long.140} {Named entity recognition with small strongly labeled and large weakly labeled data}.
\newblock In \emph{Proceedings of the 59th Annual Meeting of the Association for Computational Linguistics and the 11th International Joint Conference on Natural Language Processing (Volume 1: Long Papers)}, pages 1775--1789, Online. Association for Computational Linguistics.

\bibitem[{Levow(2006)}]{levow-2006-third}
Gina-Anne Levow. 2006.
\newblock \href {https://aclanthology.org/W06-0115} {The third international {C}hinese language processing bakeoff: Word segmentation and named entity recognition}.
\newblock In \emph{Proceedings of the Fifth {SIGHAN} Workshop on {C}hinese Language Processing}, pages 108--117, Sydney, Australia. Association for Computational Linguistics.

\bibitem[{Li et~al.(2023)Li, Zhang, Koto, Yang, Zhao, Gong, Duan, and Baldwin}]{li2023cmmlu}
Haonan Li, Yixuan Zhang, Fajri Koto, Yifei Yang, Hai Zhao, Yeyun Gong, Nan Duan, and Timothy Baldwin. 2023.
\newblock \href {http://arxiv.org/abs/2306.09212} {Cmmlu: Measuring massive multitask language understanding in chinese}.

\bibitem[{Li et~al.(2022)Li, Fei, Liu, Wu, Zhang, Teng, Ji, and Li}]{li2022unified}
Jingye Li, Hao Fei, Jiang Liu, Shengqiong Wu, Meishan Zhang, Chong Teng, Donghong Ji, and Fei Li. 2022.
\newblock Unified named entity recognition as word-word relation classification.
\newblock In \emph{Proceedings of the AAAI Conference on Artificial Intelligence}, volume~36, pages 10965--10973.

\bibitem[{Li et~al.(2016)Li, Li, He, Wang, Cao, Zhou, and Xu}]{DBLP:journals/corr/LiLHWCZX16}
Peng Li, Wei Li, Zhengyan He, Xuguang Wang, Ying Cao, Jie Zhou, and Wei Xu. 2016.
\newblock \href {http://arxiv.org/abs/1607.06275} {Dataset and neural recurrent sequence labeling model for open-domain factoid question answering}.

\bibitem[{Lou et~al.(2023)Lou, Lu, Dai, Jia, Lin, Han, Sun, and Wu}]{DBLP:journals/corr/abs-2301-03282}
Jie Lou, Yaojie Lu, Dai Dai, Wei Jia, Hongyu Lin, Xianpei Han, Le~Sun, and Hua Wu. 2023.
\newblock \href {https://doi.org/10.48550/arXiv.2301.03282} {Universal information extraction as unified semantic matching}.
\newblock \emph{AAAI}, abs/2301.03282.

\bibitem[{Lu et~al.(2022)Lu, Liu, Dai, Xiao, Lin, Han, Sun, and Wu}]{lu-etal-2022-unified}
Yaojie Lu, Qing Liu, Dai Dai, Xinyan Xiao, Hongyu Lin, Xianpei Han, Le~Sun, and Hua Wu. 2022.
\newblock \href {https://doi.org/10.18653/v1/2022.acl-long.395} {Unified structure generation for universal information extraction}.
\newblock In \emph{Proceedings of the 60th Annual Meeting of the Association for Computational Linguistics (Volume 1: Long Papers)}, pages 5755--5772, Dublin, Ireland. Association for Computational Linguistics.

\bibitem[{Robinson and Wingate(2023)}]{DBLP:conf/iclr/RobinsonW23}
Joshua Robinson and David Wingate. 2023.
\newblock \href {https://openreview.net/pdf?id=yKbprarjc5B} {Leveraging large language models for multiple choice question answering}.
\newblock In \emph{The Eleventh International Conference on Learning Representations, {ICLR} 2023, Kigali, Rwanda, May 1-5, 2023}. OpenReview.net.

\bibitem[{Roth and Yih(2004)}]{roth-yih-2004-linear}
Dan Roth and Wen-tau Yih. 2004.
\newblock \href {https://aclanthology.org/W04-2401} {A linear programming formulation for global inference in natural language tasks}.
\newblock In \emph{Proceedings of the Eighth Conference on Computational Natural Language Learning ({C}o{NLL}-2004) at {HLT}-{NAACL} 2004}, pages 1--8, Boston, Massachusetts, USA. Association for Computational Linguistics.

\bibitem[{Sheng et~al.(2021)Sheng, Guo, Yu, Li, Hei, Wang, Liu, and Xu}]{sheng-etal-2021-casee}
Jiawei Sheng, Shu Guo, Bowen Yu, Qian Li, Yiming Hei, Lihong Wang, Tingwen Liu, and Hongbo Xu. 2021.
\newblock \href {https://doi.org/10.18653/v1/2021.findings-acl.14} {{C}as{EE}: {A} joint learning framework with cascade decoding for overlapping event extraction}.
\newblock In \emph{Findings of the Association for Computational Linguistics: ACL-IJCNLP 2021}, pages 164--174, Online. Association for Computational Linguistics.

\bibitem[{Tang et~al.(2022)Tang, Xu, Zhao, Mao, Liu, Liao, and Xie}]{tang-etal-2022-unirel}
Wei Tang, Benfeng Xu, Yuyue Zhao, Zhendong Mao, Yifeng Liu, Yong Liao, and Haiyong Xie. 2022.
\newblock \href {https://aclanthology.org/2022.emnlp-main.477} {{U}ni{R}el: Unified representation and interaction for joint relational triple extraction}.
\newblock In \emph{Proceedings of the 2022 Conference on Empirical Methods in Natural Language Processing}, pages 7087--7099, Abu Dhabi, United Arab Emirates. Association for Computational Linguistics.

\bibitem[{von Werra et~al.(2020)von Werra, Belkada, Tunstall, Beeching, Thrush, and Lambert}]{vonwerra2022trl}
Leandro von Werra, Younes Belkada, Lewis Tunstall, Edward Beeching, Tristan Thrush, and Nathan Lambert. 2020.
\newblock Trl: Transformer reinforcement learning.
\newblock \url{https://github.com/lvwerra/trl}.

\bibitem[{Wang et~al.(2023)Wang, Zhou, Zu, Xia, Chen, Zhang, Zheng, Ye, Zhang, Gui, Kang, Yang, Li, and Du}]{wang2023instructuie}
Xiao Wang, Weikang Zhou, Can Zu, Han Xia, Tianze Chen, Yuansen Zhang, Rui Zheng, Junjie Ye, Qi~Zhang, Tao Gui, Jihua Kang, Jingsheng Yang, Siyuan Li, and Chunsai Du. 2023.
\newblock \href {http://arxiv.org/abs/2304.08085} {Instructuie: Multi-task instruction tuning for unified information extraction}.

\bibitem[{Wei et~al.(2023)Wei, Cui, Cheng, Wang, Zhang, Huang, Xie, Xu, Chen, Zhang, Jiang, and Han}]{wei2023zeroshot}
Xiang Wei, Xingyu Cui, Ning Cheng, Xiaobin Wang, Xin Zhang, Shen Huang, Pengjun Xie, Jinan Xu, Yufeng Chen, Meishan Zhang, Yong Jiang, and Wenjuan Han. 2023.
\newblock \href {http://arxiv.org/abs/2302.10205} {Zero-shot information extraction via chatting with chatgpt}.

\bibitem[{Xu et~al.(2018)Xu, Shen, Li, and Fu}]{DBLP:conf/cikm/XuSLF18}
Jun Xu, Siqi Shen, Dongsheng Li, and Yongquan Fu. 2018.
\newblock \href {https://doi.org/10.1145/3269206.3269272} {A network-embedding based method for author disambiguation}.
\newblock In \emph{Proceedings of the 27th {ACM} International Conference on Information and Knowledge Management, {CIKM} 2018, Torino, Italy, October 22-26, 2018}, pages 1735--1738. {ACM}.

\bibitem[{Xu and Sun(2022)}]{DBLP:journals/access/XuS22}
Jun Xu and Mengshu Sun. 2022.
\newblock \href {https://doi.org/10.1109/ACCESS.2022.3210697} {{DPNPED:} dynamic perception network for polysemous event trigger detection}.
\newblock \emph{{IEEE} Access}, 10:104801--104810.

\bibitem[{Xu et~al.(2022)Xu, Xu, Sun, Wang, and Chu}]{xu-etal-2022-extracting}
Jun Xu, Weidi Xu, Mengshu Sun, Taifeng Wang, and Wei Chu. 2022.
\newblock \href {https://aclanthology.org/2022.findings-emnlp.85} {Extracting trigger-sharing events via an event matrix}.
\newblock In \emph{Findings of the Association for Computational Linguistics: EMNLP 2022}, pages 1189--1201, Abu Dhabi, United Arab Emirates. Association for Computational Linguistics.

\bibitem[{Ye et~al.(2022)Ye, Lin, Li, and Sun}]{ye-etal-2022-packed}
Deming Ye, Yankai Lin, Peng Li, and Maosong Sun. 2022.
\newblock \href {https://doi.org/10.18653/v1/2022.acl-long.337} {Packed levitated marker for entity and relation extraction}.
\newblock In \emph{Proceedings of the 60th Annual Meeting of the Association for Computational Linguistics (Volume 1: Long Papers)}, pages 4904--4917, Dublin, Ireland. Association for Computational Linguistics.

\bibitem[{Zeng et~al.(2022)Zeng, Liu, Du, Wang, Lai, Ding, Yang, Xu, Zheng, Xia, Tam, Ma, Xue, Zhai, Chen, Zhang, Dong, and Tang}]{DBLP:journals/corr/abs-2210-02414}
Aohan Zeng, Xiao Liu, Zhengxiao Du, Zihan Wang, Hanyu Lai, Ming Ding, Zhuoyi Yang, Yifan Xu, Wendi Zheng, Xiao Xia, Weng~Lam Tam, Zixuan Ma, Yufei Xue, Jidong Zhai, Wenguang Chen, Peng Zhang, Yuxiao Dong, and Jie Tang. 2022.
\newblock \href {https://doi.org/10.48550/arXiv.2210.02414} {{GLM-130B:} an open bilingual pre-trained model}.
\newblock \emph{CoRR}, abs/2210.02414.

\bibitem[{Zhang et~al.(2023)Zhang, Cheng, Gao, and Poon}]{Zhang2022OptimizingBF}
Sheng Zhang, Hao Cheng, Jianfeng Gao, and Hoifung Poon. 2023.
\newblock Optimizing bi-encoder for named entity recognition via contrastive learning.
\newblock \emph{ICLR}.

\bibitem[{Zhang and Yang(2018)}]{zhang-yang-2018-chinese}
Yue Zhang and Jie Yang. 2018.
\newblock \href {https://doi.org/10.18653/v1/P18-1144} {{C}hinese {NER} using lattice {LSTM}}.
\newblock In \emph{Proceedings of the 56th Annual Meeting of the Association for Computational Linguistics (Volume 1: Long Papers)}, pages 1554--1564, Melbourne, Australia. Association for Computational Linguistics.

\bibitem[{Zhou et~al.(2021)Zhou, Chen, Zhao, Wu, Xu, and Li}]{Yang2021}
Yang Zhou, Yubo Chen, Jun Zhao, Yin Wu, Jiexin Xu, and Jinlong Li. 2021.
\newblock What the role is vs. what plays the role: Semi-supervised event argument extraction via dual question answering.
\newblock In \emph{Proceedings of AAAI-21}. {AAAI} Press.

\end{thebibliography}

\end{document}